\documentclass[10pt,letterpaper]{article}
\usepackage[T1]{fontenc}
\usepackage[utf8]{inputenc}
\usepackage{newtxtext,newtxmath}
\usepackage[margin=1in,top=0.85in,bottom=0.85in]{geometry}
\usepackage{microtype}
\usepackage{graphicx}
\usepackage{booktabs,array,tabularx}
\usepackage[numbers,sort&compress]{natbib}
\usepackage{enumitem}
\usepackage{caption}
\usepackage{titlesec}
\usepackage{xcolor}
\usepackage[hidelinks]{hyperref}
\usepackage{xurl}
\usepackage{needspace}

\titleformat{\section}{\large\bfseries}{\thesection}{0.6em}{}
\titlespacing*{\section}{0pt}{9pt plus 2pt minus 1pt}{4pt plus 1pt}
\setlist{nosep,leftmargin=1.4em,topsep=2pt}
\newcommand{\m}{\ensuremath{-}}
\newcommand{\passk}{pass\textasciicircum1}
\newcommand{\para}[1]{\textbf{#1}\hspace{0.4em}}

\hypersetup{pdftitle={Agent Self-Improvement with Frozen Weights: The Added Value and Limits of Failure Traces},pdfauthor={Arkid Mitra, Snehil Saluja}}

\title{\vspace{-2.2em}\Large\bfseries Agent Self-Improvement with Frozen Weights:\\ The Added Value and Limits of Failure Traces}
\author{Arkid Mitra\\ \normalsize Hopit AI\\ \normalsize\texttt{arkid@hopit.ai} \and Snehil Saluja\\ \normalsize Overlayy\\ \normalsize\texttt{snehil@overlayy.com}}
\date{\vspace{-1.2em}}

\begin{document}
\maketitle
\vspace{-0.6em}

\begin{abstract}
\noindent We evaluate whether an AI agent's harness, the notes around its model, can learn from the agent's failure traces beyond a matched search that never sees them. It can, but less than it looks. We tested an open 9-billion-parameter model with fixed weights; a larger teacher model wrote its notes. On generated customer-service decisions, where both note writers were told the target rules, notes based on the agent's mistakes improved accuracy by 4.0 percentage points over notes from the same process with the mistakes withheld. That process alone, trying and keeping better-scoring notes, reached about 60\% of the total improvement. In full customer-service conversations the same notes showed no demonstrated improvement. Separately, a strict screen of single lessons admitted two customer-service lessons, still awaiting tests on new data, and one tool-use lesson, written without seeing the agent's mistakes, which showed no demonstrated gain on new tool-use tasks. A neutral note with no procedure scored higher than the learned lessons (exploratory). A blind audit judged 146 of 240 tool-use test items defective.
\end{abstract}

\section{Introduction}

Many deployed agents are small open models whose weights stay fixed. What can still change is
the \emph{harness}: the prompt, notes, tools and control logic around the model. A growing
literature lets the agent, or a larger teacher, rewrite this harness from the agent's own traces (records of its interactions)~\citep{ace,skillopt,gepa,harness2}. We ask whether giving a teacher model access to a frozen Qwen3.5-9B agent's interaction traces helps it write notes that improve the agent's performance. We compare notes written with and without failure traces on generated retail decisions. We then compare the trace-written libraries with the standard harness in official retail conversations. Separately, we screen individual lessons; on BFCL (Berkeley Function Calling Leaderboard) multi-turn, the lessons the screen admits are also tested on held-out tasks. Every primary analysis was pre-registered.

Comparing a learned harness with its starting point mixes several sources of gain. The first is \emph{search}: a writer that proposes texts and keeps the best
ones on a development set can improve the agent without reading any failure. The second is
\emph{evidence}: the writer read the agent's failures and encoded something about them. Our comparison tests the added value of access to the episode contents within this procedure. Our control only partly separates these sources of improvement. It used the same budget limits and retained development feedback, but the actual search could differ. Its gains could therefore come from search, candidate selection, restating supplied rules, or adding text.

The question matters because recent work is sceptical. Under matched budgets, harness evolution
did not beat parallel sampling~\citep{rethinkingharness}. Failure-derived experience repairs the
source tasks of self-evolving agents but rarely transfers to unseen ones, on backbones that
include ours~\citep{repairgap,vibemem,evopath}. Some frontier agents do improve from their own
artefacts~\citep{plasticity}. On the same retail domain, ALTK-Evolve reports pass\textasciicircum4 rising from 57.5 to 67.5 on 40 held-out tasks from GPT-4.1's own traces, without a matched no-evidence control (a since-closed issue reported that its retail adapter mined test-split tasks by default)~\citep{altkevolve,altkissue}. On SkillsBench, skills an agent wrote for itself scored 42.9\%, against 45.5\% with no skills~\citep{skillinf}. Several late-2026 studies report regressions, without significance tests: all sixteen evolved agents in one study lost tasks they had previously solved~\citep{beneath}; a coding harness evolved across successive task types forgot earlier ones~\citep{continualevo}; previously solved instances regressed in 8 of 18 method--task pairs~\citep{selfimprovebench}; and memories of benign sessions led agents to skip confirmation on later high-stakes tasks~\citep{memorytrap}. Gains are also reported around frozen models from evolving prompts, tools and memory~\citep{evogroup}, and exposing the scorer's success criteria to the learner can yield gains that reflect the verifier rather than the task~\citep{criteriaexposure}. Model developers instead mostly turn agent failures into training environments for the weights~\citep{dsv41flash}, or rewrite the agent's verified runs into training data~\citep{terminalrsi}. What is still missing is a way to tell, when a loop does show a gain, whether the agent's traces produced it.

\para{Findings.} (1) On isolated, generated retail decisions, lessons that the teacher wrote from the agent's failures beat a matched search that never saw a failure by +4.0 points; the point estimate is below the 5-point gain we registered as worthwhile, although the interval includes larger gains. (2) That search alone reached about 60\% of the gain over the standard harness. (3) The same lesson libraries showed no demonstrated improvement in official $\tau^2$ retail conversations. (4) A strict screen of single lessons retained two of 12 candidate $\tau^2$ lessons and rejected four that push towards asking again; the two still need held-out confirmation. On BFCL no lesson written from traces passed the screen, and the one that passed, written without traces, did not replicate on held-out tasks. (5) On held-out BFCL a neutral note scored higher than either set of lessons (descriptive), and most instances had scoring defects.

\para{Contributions.} (i) A \emph{matched no-evidence search}: the same writer, loop and selection, with the episode contents withheld. (ii) A registered, harm-constrained \emph{gate} that admits a lesson only if randomised experiments show that it meets registered screening thresholds for gain, per-context non-inferiority within a 1-point margin, and bounded harm. (iii) A blind \emph{scorer audit}, each defect list frozen before its evaluation outcomes were opened, reported beside every BFCL estimate. (iv) Registered replications of the null result on transfer, with intervals.

\section{Setup}

\para{Agent and harness.} The agent is Qwen3.5-9B, served with vLLM with the model's built-in reasoning mode (``thinking'') off. Its weights
are never updated. The \emph{standard harness} (B1) is the benchmark's reference agent: on $\tau^2$ with the
benchmark's optional think tool added (a tool call in which the agent writes a note to itself, separate from the thinking mode), and every learned or control text goes into one notes
slot of the system prompt; on BFCL it is the official conversation without a notes message, and
a text is added as one system message. An empty text gives exactly the standard harness. An \emph{arm} is one experimental condition, specifying the harness and how its notes are assigned.

\para{Benchmarks.} \emph{$\tau^2$-bench retail}~\citep{tau2bench} is a customer-service domain
with a written policy. We use it at two levels. At the \emph{decision level}, each case is a
synthetic conversation stopped at one decision (a \emph{prefix}; for example, whether to change an address or ask
first), scored against a fixed answer: 2,997 sealed test cases over three tasks (address, variant and sequence). A \emph{wrong write} is a write tool call, such as changing an order, that is not exactly the correct one. At the
\emph{conversation level}, we run the official benchmark: 114 tasks $\times$ 4 trials, on $\tau^2$-bench v1.0.1 (commit \texttt{672227c6}); $\tau$-bench 2.0.0 (released 19--20 October 2026) is not score-comparable with 1.x~\citep{tau2v2}, so our numbers hold for the pinned version.
\emph{BFCL multi-turn}~\citep{bfcl} supplies every needed fact through the
request and the tools, so asking the user cannot replace executing the right calls. Each of its 200 \emph{roots} (task scenarios) has a Base and a Long-Context \emph{instance}. We split the roots into 50 discovery, 30 gate and 120 held-out roots (reserved for testing), giving 60 gate and 240 held-out instances; an \emph{episode} is one run of an arm on one instance.

\para{Lessons and controls.} A writer turns the agent's failures into short ``WHEN situation THEN
action'' lessons. On $\tau^2$ the writer is a larger teacher (Qwen3.8-Flash-Next-FP8; T+ in Figure~\ref{fig:main}) in a loop
of propose, check on a validation split, keep or revert, over 12 seeded runs per task; each run yields a lesson library. On BFCL the
writer is Qwen3.5-397B-A17B, reading 198 discovery episodes. We compare these lessons with two
controls.
\begin{itemize}
\item On $\tau^2$, the \emph{no-evidence search} uses the same writer, prompts, loop, selection, budget
  ceilings and development feedback, with episodes replaced by a fixed sentence; it sees exactly
  what the learned writer sees apart from the episodes. On BFCL, it matches the writer, prompts,
  loop, step count, budget and candidate screen, but receives the function documentation instead
  of episodes.
\item The \emph{neutral note} (BFCL only) is a fixed system note with no procedure: ``You can call the functions provided in this conversation. The user will make one or more requests, one turn at a time. Handle each request using the functions as needed, then reply.'' It tests
  whether adding any note changes behaviour.
\end{itemize}

\para{The gate.} Separately from these libraries, the gate screens single candidate lessons for admission: on $\tau^2$, 12 candidates formed from the teacher libraries' lessons, run on training prefixes; on BFCL, the trace and search lessons, run on the gate instances. Each candidate is switched on or
off at random across many runs, so its effect is estimated against a random background of the
other lessons. A hierarchical Bayesian model pools the estimates across contexts (task $\times$
decision type on $\tau^2$, domain family on BFCL). A lesson passes only if it has a credible gain
of more than one point, enough exposure in every context, per-context non-inferiority within
the registered 1-point margin (in every context, the probability of a loss of more than one
point is below 0.10), and bounded harm (a wrong write on $\tau^2$, a wrong final state on BFCL). On $\tau^2$ it must pass
under three named priors (sceptical, weakly informative, and slightly negative). On BFCL, lessons that pass go to a separate held-out test on the 120
unseen roots.

\para{Pre-registration.} Every primary contrast, interval method, threshold and decision rule was
written into a hash-chained registry before the data were opened. GPU experiments were run by an external collaborator from experiment kits (versioned run code and inputs) tagged at the time of each run; the registry records each
tag's commit hash. Answer keys stayed with us, and we scored the results.

\para{Scorer audit.} An external review had found score-changing defects in 8 of 15 sampled BFCL
multi-turn items~\citep{epochbfcl}. Two independent model-based reviewers labelled all 300 gate and test instances from benchmark material only, never from agent
trajectories. An instance is \emph{defective} if a plausible, natural response would have its official verdict flipped and the reviewers judge the case clear. The lists were frozen and
registered before outcomes were opened. Official scores stay primary; the audited estimate
(defective instances removed) is reported beside them and never used to choose anything.

\section{Results}

\begin{table}[t]
\centering\small
\caption{Headline registered results. Estimates are in percentage points: contrasts are first arm minus second, gate rows a lesson's estimated gain;
intervals are 95\% confidence intervals except those marked CrI (90\% credible intervals). The BFCL held-out primary is
standard harness \m{} lesson, so there a positive value favours the standard harness. Trace lessons were written from the agent's own episodes;
the no-evidence search used the same writer and procedure without them.}
\label{tab:main}
\renewcommand{\arraystretch}{1.12}
\begin{tabularx}{\linewidth}{@{}>{\raggedright\arraybackslash}X >{\raggedright\arraybackslash}p{4.6cm} >{\raggedright\arraybackslash}p{3.5cm}@{}}
\toprule
Test & Estimate & Reading \\
\midrule
$\tau^2$ decisions: trace lessons \m{} no-evidence search, accuracy (registered primary) & +4.0 [+2.3, +5.7] & passes superiority; point estimate below the 5-point worthwhile gain \\
Official $\tau^2$ retail: trace lessons \m{} standard harness, \passk & \m1.3 [\m5.9, +3.3] & no demonstrated improvement \\
$\tau^2$ gate: ``address change first'' / ``list only the first change'' / both together & +4.7 [+4.0, +5.4] / +2.7 [+2.0, +3.4] / +7.4 [+6.3, +8.4] (CrI) & passes the screen; no held-out test was run \\
BFCL gate: best lesson (from the no-evidence search); 0 of 8 trace lessons pass & +4.9 [+1.9, +8.0]; audited +3.9 [\m0.4, +8.3] (both CrI) & passes the screen \\
BFCL held-out: standard harness \m{} that lesson (positive favours the standard harness) & +0.31 [\m3.12, +3.65], Holm $p$ 1.000 & null; did not replicate \\
BFCL held-out: neutral note \m{} standard harness (descriptive) & +4.5 [+0.7, +8.4] & registered descriptive secondary; higher observed score, unadjusted for multiple tests; no confirmatory claim \\
\midrule
Scorer audit: clearly defective BFCL instances & 146 of 240 held-out; 42 of 60 gate ($\kappa$ 0.84) & qualifies every BFCL estimate \\
\bottomrule
\end{tabularx}
\end{table}

\begin{figure}[t]
\centering
\includegraphics[width=\linewidth]{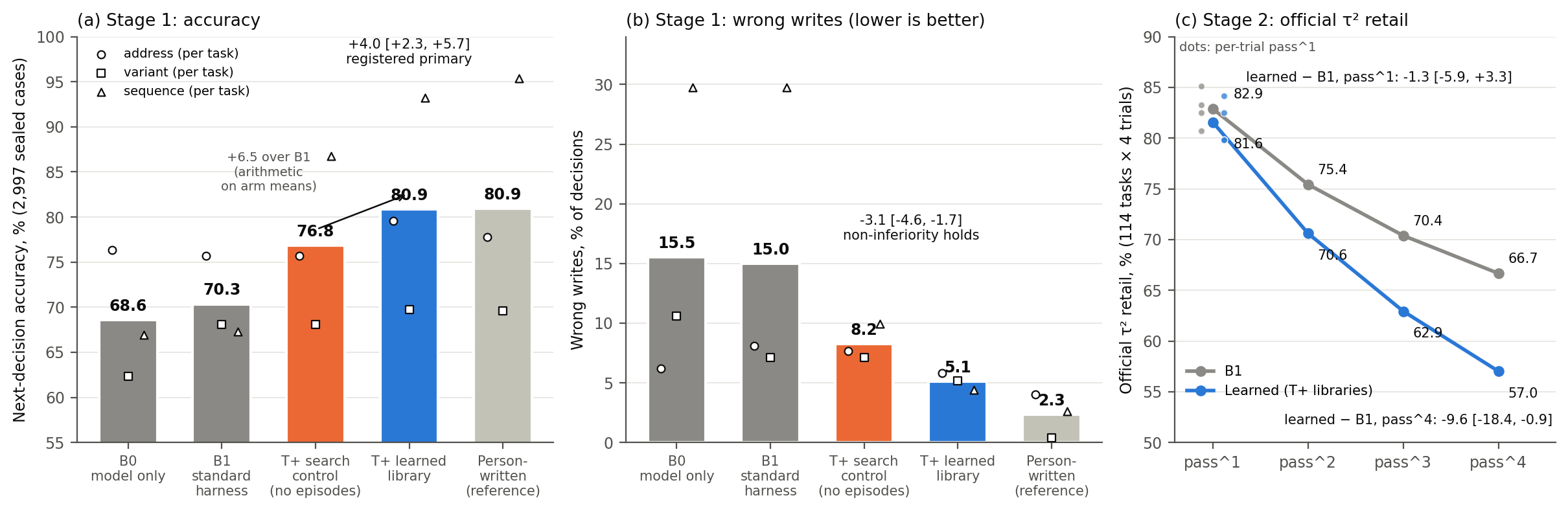}
\caption{(a) Decision-level (``Stage 1'') accuracy on 2,997 sealed $\tau^2$ retail cases: model only (B0),
standard harness (B1), no-evidence search (``T+ search control''), trace lessons (``T+ learned
library''; T+ is the teacher writer) and a person-written reference. Markers are per-task values for the address, variant and sequence tasks. (b) Wrong writes (write calls that are not exactly the correct one) for the
same arms. (c) Official $\tau^2$ retail conversations (``Stage 2''; 114 tasks $\times$ 4 trials): pass\textasciicircum$k$
for the standard harness and the trace lessons. pass\textasciicircum$k$ is the chance that all $k$ trials of a task succeed, so it cannot increase as $k$ grows (unlike pass@$k$, where any one success counts). The learned arm uses a different lesson library in
each trial while the standard harness uses one prompt, so for $k>1$ its pass\textasciicircum$k$ also
reflects variation across libraries; the pass\textasciicircum4 gap is exploratory.}
\label{fig:main}
\end{figure}

\para{Decision level: the traces add a little.} On the 2,997 sealed $\tau^2$ decisions, the
standard harness scored 70.3\%, the no-evidence search 76.8\% and the trace-written lessons 80.9\%
(Figure~\ref{fig:main}a, Table~\ref{tab:main}). The registered contrast, lessons minus search, is +4.0 [+2.3, +5.7]. It passes the superiority test; its point estimate falls below the 5-point gain we registered as worthwhile, though the interval includes larger gains.
Relative to the search, wrong writes fell by 3.1 points [\m4.6, \m1.7]. The lessons improve on the standard harness by +10.5 points [+8.5, +12.5]. The control achieves 6.5 of those 10.5 points, about 60\%, without traces. That entire control gain occurs on the task whose rule is stated in the writer's task description. 

\para{Conversations: no demonstrated improvement.} In official $\tau^2$ retail (114 tasks $\times$ 4 trials),
\passk{} (the single-trial task success rate) was 82.9\% for the standard harness and 81.6\% with the lessons: \m1.3 [\m5.9, +3.3]. On
the 40 tasks we had never used in development it was \m0.6 [\m6.9, +6.3]. The lessons changed the targeted behaviour: the number of conversations in which the agent changed items before changing the address fell from 6 to 0.
These behaviours can affect which final states are reachable, and $\tau^2$ rewards the final state.

\para{The $\tau^2$ gate admits two lessons, as a screen.} From 12 candidate teacher lessons, run on
20,400 decisions over 1,200 prefixes, two passed every check under every prior: ``change the
address first'' (+4.7 [+4.0, +5.4]) and ``list only the first change'' (+2.7 [+2.0, +3.4]);
together +7.4 [+6.3, +8.4] (all three are 90\% CrIs). Four lessons that push towards asking or confirming again were
rejected, because each credibly hurt at least one context. These are estimates on the data used
to select them. No held-out test was run, and the first lesson states the rule that the
no-evidence search also wrote almost word for word.

\para{BFCL gate: only the search passes.} On 60 gate instances (2,820 episodes), none of the 8
trace-written lessons passed; every estimate lay within $\pm$2.5 points. The one lesson that
passed came from the no-evidence search: ``when the current directory is unknown, call
\texttt{pwd}'', at +4.9 [+1.9, +8.0] (CrI). On the 18 instances the audit judged clean, its estimate fell to +3.9 [\m0.4, +8.3] (CrI), an interval that includes zero.

\para{BFCL held-out: no demonstrated gain.} On 120 held-out roots (240 instances, 5,760 episodes), the registered difference
between the standard harness and the search lesson was +0.31 [\m3.12, +3.65] (Holm-adjusted $p$ 1.000).
As the lesson's gain this is \m0.31 [\m3.65, +3.12], against +4.9 at the gate, which lies outside the interval; smaller positive effects remain compatible. The other registered contrast, standard harness minus all eight trace lessons run without the gate, was \m0.52 [\m4.38, +3.33] (Holm $p$ 1.000). In registered descriptive secondary analyses, not corrected for multiple tests, the neutral note gained +4.5 [+0.7, +8.4] over the standard harness, more than either set of lessons; on clean instances this shrank to +3.4 [\m1.9, +9.1]. The secondary analyses were registered, but the observation that the neutral note outperformed both lesson sets was not.

\para{Scorer defects.} The audit judged 146 of 240 held-out instances (60.8\% [54.5, 66.8]) and 42
of 60 gate instances clearly defective, almost all because a correct response is scored wrong.
Reviewer agreement was $\kappa = 0.84$. On clean held-out instances both learned-lesson arms sat
about 6 points \emph{below} the standard harness: standard harness \m{} search lesson +6.0
[+0.2, +12.0] (Holm $p$ 0.149), the primary's sign but much larger than +0.31; standard harness
\m{} the eight trace lessons +6.2 [\m0.5, +13.2] (Holm $p$ 0.153), the reverse of \m0.52. By the
registered rule the primary stands.

\section{What this means}

\para{Access to mistakes helped on generated decisions.} Access to the agent's mistakes improved decision accuracy within this procedure; the point estimate was below the registered worthwhile gain, although the interval includes larger gains. No improvement from learned lessons was demonstrated in full conversations or on held-out BFCL tasks. This agrees with other studies of small open
models~\citep{repairgap,vibemem,lifeharness,recuris}. Our addition is the controlled comparison: most of the decision-level gain came without reading failures.

\para{Screens are not certifications.} The two admitted $\tau^2$ lessons remain untested in conversations, and on BFCL the admitted lesson's gate gain was not confirmed on held-out tasks. A selected winner should shrink on fresh data, and a gate on one split cannot tell a small true effect from selection on noise. Noise is large: across four identical trials of published $\tau^2$ retail configurations, 33--55\% of task verdicts flipped~\citep{tau2noise}. 

\para{Some note effects may be non-specific.} Because the neutral note scored higher than either set of lessons on held-out BFCL (descriptively), a learned note should be compared with a neutral note, not only with no note.

\para{Benchmark scorers need audits.} Most BFCL instances had clear defects, and the audit changed how strongly results can be read: the gate winner lost its credible gain on clean instances, and on clean held-out instances both learned-lesson arms sat about 6 points below the standard harness. Freezing it before outcomes kept it from steering results.

\para{Limitations.} We study one frozen 9B model, with thinking off, on two benchmarks. On
$\tau^2$ the writers' task descriptions stated the target rules, so the decision-level result
measures whether a teacher can phrase a known rule so that the 9B follows it, not whether it can
discover a rule. On BFCL the search writer saw documentation the trace writer did not. The
decision cases are snapshots: the agent cannot explore or recover. Several deviations in the collaborator-run experiments are disclosed in full in the extended report. The registered teacher was
replaced before any result; the $\tau^2$ gate data were re-run in full after the first run's
folder was lost; in that run, read calls were answered by the original database (visible in
0.27\% of rows; removing them leaves the admitted lessons unchanged); BFCL used sampled decoding
(temperature 0.7, not the official 0.001); and a scorer serialisation fix was applied after the
run, with verdicts byte-identical wherever they could be compared.

\para{What we do not claim.}
\begin{itemize}
\item That the lessons have no effect on full conversations or held-out BFCL tasks: the intervals allow a few points either way.
\item That the two admitted $\tau^2$ lessons are certified, or that their gain comes from the
  traces.
\item That these results extend beyond one agent, two benchmarks and one teacher family.
\end{itemize}

\section{Conclusion}

For a frozen Qwen3.5-9B agent, giving the teacher access to the agent's mistakes helped: its notes improved accuracy on generated decisions by +4.0 points over the matched search without episode contents. Both writers already had the target rules. But no improvement was demonstrated
on full episodes, and without the matched search the decision-level gain would have looked 2.6
times larger. We expect the methods to outlast the answer: report a matched no-evidence search, compare
notes with a neutral note, confirm gate winners on held-out data, and audit the scorer blind,
before the comparison.

\para{Data and code availability.} Analysis code, this study's pre-registration records, the
frozen defect lists and derived outcome tables are public at \url{https://github.com/hopit-ai/harness-mistakes-artifacts}
(PolyForm Noncommercial licence); the experiment kits and other materials are available from the
authors on request (arkid@hopit.ai). An extended version with all registered analyses is available
from the authors.

\bibliographystyle{unsrtnat}
\bibliography{refs}

\end{document}